\documentclass[sigconf]{acmart}

\AtBeginDocument{%
  \providecommand\BibTeX{{%
    \normalfont B\kern-0.5em{\scshape i\kern-0.25em b}\kern-0.8em\TeX}}}

\usepackage{balance}
\usepackage{fancyhdr}
\newtheorem{assumption}{Assumption}[section]
\newcommand{\independent}{\protect\mathpalette{\protect\independenT}{\perp}}
\def\independenT#1#2{\mathrel{\rlap{$#1#2$}\mkern2mu{#1#2}}}

\begin{document}
\fancyhead{}
\title{Graph Infomax Adversarial Learning for Treatment Effect Estimation with Networked Observational Data}

\author{Zhixuan Chu}
\email{zhixuan.chu@uga.edu}
\affiliation{
   \institution{University of Georgia}
   \city{Athens}
   \state{Georgia}
}

\author{Stephen L. Rathbun}
\email{rathbun@uga.edu}
\affiliation{
   \institution{University of Georgia}
   \city{Athens}
   \state{Georgia}
}

\author{Sheng Li}
\email{sheng.li@uga.edu}
\affiliation{%
   \institution{University of Georgia}
   \city{Athens}
   \state{Georgia}
}

\renewcommand{\shortauthors}{Chu, et al.}

\begin{abstract}
Treatment effect estimation from observational data is a critical research topic across many domains. The foremost challenge in treatment effect estimation is how to capture hidden confounders. Recently, the growing availability of networked observational data offers a new opportunity to deal with the issue of hidden confounders. Unlike networked data in traditional graph learning tasks, such as node classification and link detection, the networked data under the causal inference problem has its particularity, i.e., imbalanced network structure. In this paper, we propose a Graph Infomax Adversarial Learning (GIAL) model for treatment effect estimation, which makes full use of the network structure to capture more information by recognizing the imbalance in network structure. We evaluate the performance of our GIAL model on two benchmark datasets, and the results demonstrate superiority over the state-of-the-art methods.

\end{abstract}

\maketitle

\section{Introduction}
A further understanding of causality beyond observational data is critical across many domains including statistics, computer science, education, public policy, economics, and health care. Although randomized controlled trials (RCT) are usually considered as the gold standard for causal inference, estimating causal effects from observational data has received growing attention owing to the increasing availability of data and the low costs compared to RCT.

When estimating treatment effects from observational data, we face two major issues, i.e., missing counterfactual outcomes and treatment selection bias. The foremost challenge for solving these two issues is the existence of confounders, which are the variables that affect both treatment assignment and outcome. Unlike RCT, treatments are typically not assigned at random in observational data. Due to the confounders, subjects would have a preference for a certain treatment option, which leads to a bias of the distribution for the confounders among different treatment options. This phenomenon exacerbates the difficulty of counterfactual outcome estimation. For most of existing methods~\cite{hill2011bayesian,chu2020matching, li2016matching, li2017matching, shalit2017estimating, wager2018estimation, yao2018representation, alaa2017bayesian, yao2019estimation}, the \emph{strong ignorability} assumption is the most important prerequisite. It assumes given covariates, the treatment assignment is independent of the potential outcomes and for any value of covariates, treatment assignment is not deterministic. Strong ignorability is also known as the \emph{no unmeasured confounders} assumption. This assumption requires that all the confounders be observed and sufficient to characterize the treatment assignment mechanism. Moreover, strong ignorability is a sufficient condition for the individual treatment effect (ITE) function to be identifiable~\cite{imbens2009recent}. 

However, due to the fact that identifying all of the confounders is impossible in practice, the strong ignorability assumption is usually untenable. By leveraging big data, it becomes possible to find a proxy for the hidden confounders. Network information, which serves as an efficient structured representation of non-regular data, is ubiquitous in the real world. Advanced by the powerful representation capabilities of various graph neural networks, networked data has recently received increasing attention~\cite{kipf2016semi,velivckovic2017graph,velickovic2019deep,jiang2019censnet}. Besides, it can be used to help recognize the patterns of hidden confounders. A network deconfounder~\cite{guo2019learning} is proposed to recognize hidden confounders by combining the graph convolutional networks~\cite{kipf2016semi} and counterfactual regression~\cite{shalit2017estimating}. 

\begin{figure}[t]
    \centering
    \includegraphics[width=0.3\textwidth]{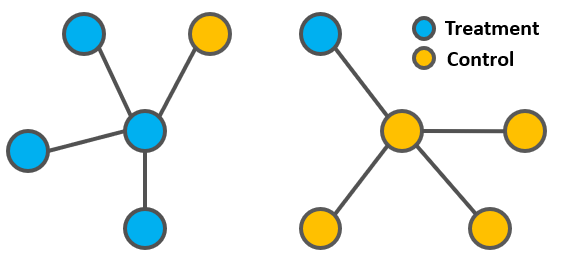}
    \vspace{-2mm}
    \caption{Example of the imbalance of network structure.}
    \label{exmaple}
    \vspace{-2mm}
\end{figure}

\begin{figure*}[t]
    \centering
    \includegraphics[width=0.85\textwidth]{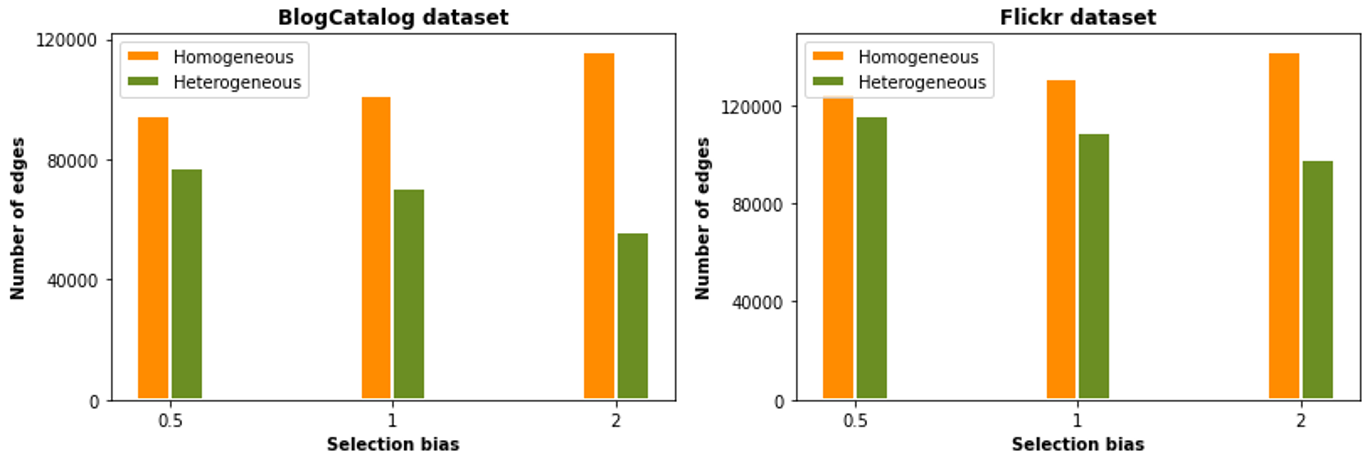}
    \vspace{-2mm}
    \caption{Under the assumption that each node has the same possibility to be connected with another node regardless of node's treatment assignment, for $n$ nodes, there should be $\frac{n^2}{4}-\frac{n}{2}$ homogeneous edges (that link the nodes in the same group, i.e., treatment-treatment or control-control) and $\frac{n^2}{4}$ heterogeneous edges (that link the nodes in different groups, i.e., treatment-control). The number of heterogeneous edges should be greater than that of homogeneous edges. However, in the benchmarks of causal inference with networked data (\textit{BlogCatalog} and \textit{Flickr}), the homogeneous edges are consistently greater than heterogeneous edges for both datasets. Besides, as the selection bias increases, the difference between homogeneous and heterogeneous edges gets larger. This result totally agrees with our expectation that, in the causal inference problem, the network structure is imbalanced. The relationship is more likely to appear among people who are in the same group.}
    \label{imbalance}
\end{figure*}

The networked observational data consists of two components, node features and network structures. Due to the confounding bias in causal inference problem, the imbalance not only exists in distributions of feature variables in treatment and control groups but also in network structures. For example, in social networks, the links are more likely to appear among more similar people, so the subjects are more likely to follow other subjects in the same group as shown in Fig.~\ref{exmaple}, which will aggravate the imbalance in the representation space learned by graph neural networks. Fig.~\ref{imbalance} shows the existence of imbalanced network structures in the benchmarks of causal inference with networked data (\textit{BlogCatalog} and \textit{Flickr}). Unlike the networked data in traditional graph learning tasks, such as node classification and link detection, the networked data under the causal inference problem has its particularity, i.e., imbalanced network structure. For most existing work on networked observational data, they did not consider this peculiarity of graph structure under causal inference settings. Directly applying graph neural networks designed for traditional graph learning tasks cannot capture all of the information from imbalanced networked data. 

To fully exploit the information in the networked data with the imbalanced network structure, we propose a Graph Infomax Adversarial Learning method (GIAL) to estimate the treatment effects from networked observational data.  In our model, structure mutual information is maximized to help graph neural networks to extract a representation space, which best represents observed and hidden confounders from the networked data with the imbalanced structure. Also, adversarial learning is applied to balance the learned representation distributions of treatment and control groups and to generate the potential outcomes for each unit across two groups. Overall, GIAL can make full use of network structure to recognize patterns of hidden confounders, which has been validated by extensive experiments on benchmark datasets.

\section{Background}
Suppose that the observational data contain $n$ units and each unit received one of two or more treatments. Let $t_i$ denote the treatment assignment for unit $i$; $i=1,...,n$. For binary treatments, $t_i = 1$ is for the treatment group, and $t_i=0$ for the control group. The outcome for unit $i$ is denoted by $Y_{t}^i$ when treatment $t$ is applied to unit $i$; that is, $Y_{1}^i$ is the potential outcome of unit $i$ in the treatment group and $Y_0^i$ is the potential outcome of unit $i$ in the control group. For observational data, only one of the potential outcomes is observed according to the actual treatment assignment of unit $i$. The observed outcome is called the factual outcome, and the remaining unobserved potential outcomes are called counterfactual outcomes. Let $X \in \mathbb{R}^d$ denote all observed variables of a unit. 

Let $\mathcal{G}(\mathcal{V},\mathcal{E})$ denote an undirected graph, where $\mathcal{V}$ represents $n$ nodes in $\mathcal{G}$ and $\mathcal{E}$ is a set of edges between nodes. According to the adjacency relationships in $\mathcal{E}$, the corresponding adjacent matrix $A \in \mathbb{R}^{n \times n}$ of the graph $\mathcal{G}$ can be defined as follows. If $(v_i,v_j) \in \mathcal{E}$, $A_{ij} = 1$, otherwise $A_{ij}=0$. When edges have different weights, $A_{ij}$ can be assigned to a real value. 

In this paper, we explore the observational data as networks. In particular, the graph $\mathcal{G}$ is the networked observational data. Every node in $\mathcal{V}$ is one unit in observational data, an edge in $\mathcal{V}$ describes the relationship between a pair of units, and adjacent matrix $A$ represents the whole network structure. Therefore, the observational data can be denoted as $(\{x_i, t_i, y_i\}_{i=1}^n,A)$. We follow the potential outcome framework for estimating treatment effects ~\cite{rubin1974estimating}. The individual treatment effect (ITE) for unit $i$ is the difference between the potential treated and control outcomes, which is defined as: $ \text{ITE}_i = Y_1^i - Y_0^i, \quad (i=1,...,n).$

The average treatment effect (ATE) is the difference between the mean potential treated and control outcomes, which is defined as $\text{ATE}=\frac{1}{n}\sum_{i=1}^{n}(Y_1^i - Y_0^i), \quad (i=1,...,n).$ The success of the potential outcome framework is based on the strong ignorability assumption, which ensures that the treatment effect can be identified~\cite{imbens2015causal, yao2020survey}.

\begin{assumption} \textbf{Strong Ignorability}: 
Given covariates $X$, treatment assignment $T$ is independent of the potential outcomes, i.e., $(Y_1, Y_0) \independent T | X$ and for any value of $\,X$, treatment assignment is not deterministic, i.e.,$P(T = t | X = x) > 0$, for all $t$ and $x$.
\end{assumption}

In our model, we relax the strong ignorability and allow the existence of hidden confounders. We aim to use network structure information to recognize the hidden confounders and then estimate treatment effects based on the learned confounder representations.

\section{The Proposed Framework}
\subsection{Motivation}

The foremost challenge of causal inference from observational data is how to recognize hidden confounders. Recently, leveraging the powerful representation capabilities of various graph neural networks, network structures can be utilized to help recognize the patterns of hidden confounders in networked observational data. 

Due to the particularity of the causal inference problem, the networked data in causal inference is different from that in traditional graph learning tasks such as node classification and link detection. As network information is incorporated into the model, we face a new imbalance issue,i.e., imbalance of network structure in addition to the imbalance of observed covariate distributions. A link has a larger probability of appearing between two more similar people. It implies that one unit is more likely to be connected to other units in the same group. Therefore, directly applying traditional graph learning methods to learn the representation of networked data could not fully exploit the useful information for causal inference. 

It is essential to design a new method that can capture the representation of hidden confounders implied from the imbalanced network structure and observed confounders that exist in the covariates simultaneously. To solve this problem, we propose the Graph Infomax Adversarial Learning method (GIAL) to estimate the treatment effects from the networked observational data, which can recognize patterns of hidden confounders from imbalanced network structure.

\begin{figure*}[h]
    \centering
    \includegraphics[width=0.8\textwidth]{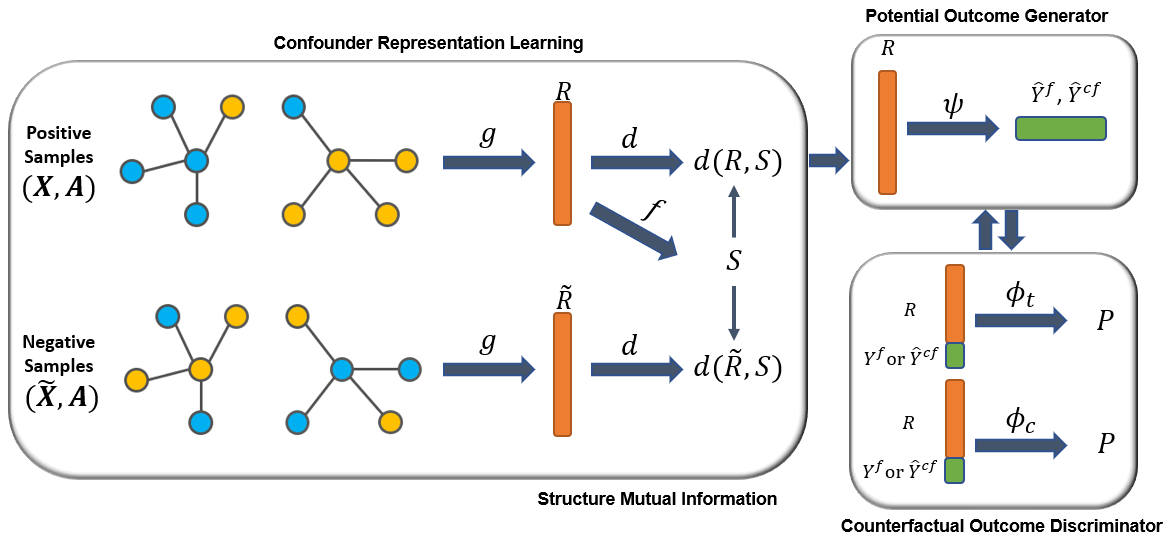}
    \vspace{-2mm}
    \caption{Framework of our Graph Infomax Adversarial Learning method (GIAL). Graph neural networks and structure mutual information are utilized to learn the representations of hidden confounders and observed confounders. Then the potential outcome generator is applied to infer the potential outcomes of units across treatment and control groups based on the learned representation space and treatment assignment. At the same time, the counterfactual outcome discriminator is incorporated to remove the imbalance in the learned representations of treatment and control groups.}
    \vspace{-2mm}
    \label{framework}
\end{figure*}
\subsection{Model Architecture}
As shown in Fig.~\ref{framework}, our GIAL consists of four main components, i.e., confounder representation learning, structure mutual information maximization, potential outcome generator, and counterfactual outcome discriminator. Firstly, we utilize the graph neural network and structure mutual information to learn the representations of hidden confounders and observed confounders, by mapping the feature covariates and network structure simultaneously into a representation space. Then the potential outcome generator is applied to infer the potential outcomes of units across treatment and control groups based on the learned representation space and treatment assignment. At the same time, the counterfactual outcome discriminator is incorporated to remove the imbalance in the learned representations of treatment and control groups, and thus it improves the prediction accuracy of potential outcomes inferred in the outcome generator by playing a minimax game. In the following, we present the details of each component.

\textbf{Confounder Representation Learning.}
Based on the graph $\mathcal{G}(\mathcal{V},\mathcal{E})$, our goal is to learn the representation of confounders by a function $g : X \times A \rightarrow R, R \in \mathbb{R}^d$, which is parameterized by a graph neural network. To better capture information resided in the networked data, we separately adopt two powerful graph neural network methods, i.e., the graph convolutional network (GCN) ~\cite{kipf2016semi} and graph attention network layers (GAT) ~\cite{velivckovic2017graph}, to learn the representation space. For these two models, their effectiveness of the learned representations has been verified in various graph learning tasks. The major difference between GCN and GAT is how the information from the one-hop neighborhood is aggregated. For GCN, a graph convolution operation is used to produce the normalized sum of the node features of neighbors. GAT introduces the attention mechanism to better quantify the importance of each edge. Here, we want to find out which model is better to unravel patterns of hidden confounders from the networked data with imbalanced covariate and imbalanced network structure.

For the graph convolutional network (GCN) model, the representation learning function $g : X \times A \rightarrow R$ is parameterized with the following layer-wise propagation rule:
\begin{equation}
r^{(l+1)}=\sigma (\Tilde{D}^{-\frac{1}{2}}\Tilde{A}\Tilde{D}^{-\frac{1}{2}}r^{(l)}W^{(l)}),
\end{equation}
where $\Tilde{A}=A+I_n$ is the adjacency matrix of graph $\mathcal{G}(\mathcal{V},\mathcal{E})$ with inserted self-loops, i.e., the identity matrix $I_n$. $\Tilde{D}$ is its corresponding degree matrix, i.e., $\Tilde{D}_{ii} = \sum_{j}\Tilde{A}_{ij}$ and $W^{(l)}$ is a layer-specific trainable weight matrix. $\sigma(\cdot)$ denotes an activation function and here we apply the parametric ReLU (PReLU) function ~\cite{he2015delving}. A number of GCN layers can be stacked to approximate the function $g : X \times A \rightarrow R$.

For the graph attention network (GAT) model, the representation of confounder for the $i$-th node is a function of its covariates and receptive field. Here, we define the $i$-th node $v_i$ and its one-hop neighbor nodes as the receptive field $\mathcal{N}(v_i)$. The representation learning function $g : X \times A \rightarrow R$ is parameterized with the following equation:
\begin{equation}
r^{(l+1)}_i=\sigma\left(\sum_{j\in \mathcal{N}(i)} {\alpha^{(l)}_{ij} W^{(l)}r^{(l)}_j}\right),
\label{eq-gat}
\end{equation}
where $W^{(l)}$ is the learnable weight matrix and $W^{(l)}r^{(l)}_j$ is a linear transformation of the lower layer representation $r^{(l)}_j$. $\sigma(\cdot)$ is the activation function for nonlinearity. In Eq.~\eqref{eq-gat}, the representation of the $i$-th node and its neighbors are aggregated together, scaled by the normalized attention scores $\alpha^{(l)}_{ij}$.

\begin{equation}
\alpha^{(l)}_{ij}=\frac{\exp(\text{LeakyReLU}( {a^{(l)}}^T(W^{(l)}r^{(l)}_i||W^{(l)}r^{(l)}_j)))}{\sum_{k\in \mathcal{N}(i)}^{}\exp(\text{LeakyReLU}( {a^{(l)}}^T(W^{(l)}r^{(l)}_i||W^{(l)}r^{(l)}_k)))},
\end{equation}
where softmax is used to normalize the attention scores on each node's incoming edges. The pair-wise attention score between two neighbors is calculated by $\text{LeakyReLU}(a^{(l)^T}(W^{(l)}r^{(l)}_i||W^{(l)}r^{(l)}_j))$. Here, it first concatenates the linear transformation of the lower layer representations for two nodes, i.e., $W^{(l)}r^{(l)}_i||W^{(l)}r^{(l)}_j$,  where $||$ denotes concatenation, and then it takes a dot product of itself and a learnable weight vector $a^{(l)}$. Finally, the LeakyReLU function is applied.

To stabilize the learning process, a multi-head attention mechanism is employed. We compute multiple different attention maps and finally aggregate all the learned representations. In particular, $K$ independent attention mechanisms execute the transformation of  Eq.~\eqref{eq-gat}, and then their outputs are merged in two ways:
\begin{equation}
\text{concatenation}: r^{(l+1)}_{i} =||_{k=1}^{K}\sigma\left(\sum_{j\in \mathcal{N}(i)}\alpha_{ij}^{k}W^{k}r^{(l)}_{j}\right)
\end{equation}

or

\begin{equation}
\text{average}: h_{i}^{(l+1)}=\sigma\left(\frac{1}{K}\sum_{k=1}^{K}\sum_{j\in\mathcal{N}(i)}\alpha_{ij}^{k}W^{k}h^{(l)}_{j}\right)
\end{equation}

When performing the multi-head attention on the final layer of the network, concatenation is no longer sensible. Thus, we use the concatenation for intermediary layers and the average for the final layer. An arbitrary number of GAT layers can be stacked to approximate the function $g: X \times A \rightarrow R$.

\textbf{Structure Mutual Information Maximization.}
Inspired by a recent successful unsupervised graph learning method ~\cite{velickovic2019deep}, we maximize structure mutual information to capture the imbalanced graph structure with respect to treatment and control nodes in the networked observational data. We aim to learn representations that can capture the imbalanced structure of the entire graph. Specifically, we utilize a structure summary function, $f : R \rightarrow S, S \in \mathbb{R}^d$, to summarize the learned representation into an entire graph structure representation, i.e., $S=f(g(X,A))$. From the observations in empirical evaluations, the structure summary function could be defined as $s = \sigma(\frac{1}{n}\sum_{i=1}^{n}r_i)$ to best capture the entire graph structure, where $\sigma$ is the logistic sigmoid activation function. 

Here, our purpose is to learn a representation vector, which can capture the entire graph structure encoded by the graph structure summary vector $s$ and also reflect the abnormal imbalance in the graph structure. Therefore, we aim at maximizing the mutual information between the learned representation vector $r_i$ and the structure summary vector $s$.  

Mutual information is a fundamental quantity for measuring the relationship between random variables. For example, the dependence of two random variables $W$ and $Z$ is quantified by mutual information as~\cite{belghazi2018mutual}:
\begin{equation}
 I(W;Z)=\int_{\mathcal{W}\times\mathcal{Z}}{\text{log}\frac{d\mathbb{P}_{WZ}}{d\mathbb{P}_W\otimes\mathbb{P}_Z}d\mathbb{P}_{WZ}},
\end{equation}
where $\mathbb{P}_{WZ}$ is the joint probability distribution, and $\mathbb{P}_W=\int_{\mathcal{W}}d\mathbb{P}_{WZ}$ and $\mathbb{P}_Z=\int_{\mathcal{Z}}d\mathbb{P}_{WZ}$ are the marginals. However, mutual information has historically been difficult to compute. From the viewpoint of Shannon
information theory, mutual information can be estimated as Kullback-Leibler divergence: 
\begin{equation}
 I(W;Z)=H(W)-H(W|Z)=D_{KL}(\mathbb{P}_{WZ} || \mathbb{P}_W\otimes\mathbb{P}_Z).
\end{equation}

Actually, in our model, it is unnecessary to use the exact KL-based formulation of MI, as we only want to maximize the mutual information between representation vector $r_i$ and structure summary vector $s$. A simple and stable alternative based on the Jensen-Shannon divergence (JSD) can be utilized. Thus, we follow the intuitions from deep infomax~\cite{hjelm2018learning} and deep graph infomax ~\cite{velickovic2019deep} to maximize the mutual information.

To act as an agent for maximizing the mutual information, one discriminator $d: R \times S \rightarrow P, P \in \mathbb{R}$ is employed. The discriminator is formulated by a simple bilinear scoring function with nonlinear activation: $d(r_i,s)=\sigma({r_i}^TWs)$, which estimates the probability of the $i$-th node representation contained within the graph structure summary $s$. $W$ is a learnable scoring matrix. 

To implement the discriminator, we also need to create the negative samples compared with original samples and then use the discriminator to distinguish which one is from positive samples (original networked data) and which one is from the negative samples (created fake networked data), such that the original graph structure information could be correctly captured. The choice of the negative sampling procedure will govern the specific kinds of structural information that is desirable to be captured ~\cite{velickovic2019deep}. Here, we focus on the imbalance between the edges that link nodes in the same group and those that link nodes in the different groups, i.e., treatment unit to treatment unit, treatment unit to control unit, and control unit to control unit. Therefore, our discriminator is designed to force the representations to capture this imbalanced structure by creating negative samples where the original adjacency matrix $A$ is preserved, whereas the negative samples $\Tilde{X}$ are obtained by the row-wise shuffling of $X$. That is, the created fake networked data consists of the same nodes as the original graph, but they are located in different places in the same structure. Thus, the nodes at both ends of the edges may change the treatment choices, e.g., from treatment to control, from control to treatment, or remain unchanged. Then we also conduct the confounder representation learning for the created fake networked data $(\Tilde{X}, A)$ to get the $\Tilde{r}_i$. With the proposed discriminator, we could have $d(r_i,s)$ and $d(\Tilde{r}_i,s)$, which indicate the probabilities of containing the representations of the $i$-th positive sample and negative sample in the graph structure summary, respectively.

We optimize the discriminator to maximize mutual information between $r_i$ and $s$ based on the Jensen Shannon divergence via a noise-contrastive type objective with a standard binary cross-entropy (BCE) loss~\cite{velickovic2019deep,hjelm2018learning}:
\begin{equation}
\mathcal{L}_m = \frac{1}{2n}\Big(\sum_{i=1}^n \mathbb{E}_{(X,A)}[\text{log} \, d(r_i,s)]+\sum_{j=1}^n \mathbb{E}_{(\Tilde{X},A)}[\text{log} \, (1-d(\Tilde{r}_i,s))]\Big). 
\label{eq: mutual information}
\end{equation}

\textbf{Potential Outcome Generator.} 
So far, we have learned the representation space of confounders from networked data with the imbalanced network structure and imbalanced covariates. The function $\Psi: R \times T \rightarrow Y$ maps the representation of hidden confounders and observed confounders as well as a treatment to the corresponding potential outcome, which is parameterized by a feed-forward deep neural network with multiple hidden layers and non-linear activation functions. The function $\Psi: R \times T \rightarrow Y$ uses representations and treatment options as inputs to predict potential outcomes. The output of $\Psi$ estimates potential outcomes across treatment and control groups, including the estimated factual outcome $\hat{y}^f$ and the estimated counterfactual outcomes $\hat{y}^{cf}$. The factual outcomes $y^f$ are used to minimize the loss of prediction $\hat{y}^f$. We aim to minimize the mean squared error in predicting factual outcomes:
\begin{equation}
\mathcal{L}_\Psi = \frac{1}{n}\sum_{i=1}^{N}(\hat{y}^f_i-y^f_i)^2,
\label{eq: factual}
\end{equation}
where $\hat{y}_i=\Psi(r_i,t_i)$ denotes the inferred observed outcome of unit $i$ corresponding to the factual treatment $t_i$. 

\textbf{Counterfactual Outcome Discriminator.} The counterfactual outcome discriminator is intended to remove the imbalance of confounder representations between treatment and control groups, and thus it could improve the prediction accuracy of potential outcomes inferred by the outcome generator. We define the counterfactual outcome discriminator as $\Phi: R \times T \times (Y^f \text{or} \ \hat{Y}^{cf}) \rightarrow P$, where $P$ is the discriminator's judgement, i.e., probability that this outcome for unit $i$ given $R$ and $T$ is factual outcome. $P$ is defined as: 
\begin{equation}
P=  
  \begin{cases} 
   P(\text{judges} \ y^f \text{as factual} |x,t) \, \text{if} \ t \ \text{is factual treatment choice}  \\
   P(\text{judges} \ \hat{y}^{cf} \text{as factual} |x,t)\, \text{if} \ t \ \text{is not factual treatment choice}. \\
  \end{cases}
  \label{Eqn: p}
\end{equation}

To improve the accuracy of prediction and avoid risk of losing the influence of treatment $t$ and potential outcomes $(y^f\text{or} \ \hat{y}^{cf})$ due to high dimensional representation vector, we adopt separate head networks for treatment and control groups ~\cite{shalit2017estimating}. Besides, to improve the influence of $(y^f,\hat{y}^{cf})$ in the discriminator, we add $(y^f \text{or} \ \hat{y}^{cf})$ into each layer of the neural network, repetitively.

The discriminator deals with a binary classification task, which assigns one label (i.e., factual outcome or counterfactual outcome) to the vector concatenating the representation vector $r$ and potential outcome $(y^f \text{or} \ \hat{y}^{cf})$ under the treatment head network and control head network, respectively. Thus, the loss of discrimination is measured by the cross-entropy with truth probability, where $P^{\text{truth}}=1$ if $y^f$ is input, and $P^{\text{truth}}=0$ if $\hat{y}^{cf}$ is input. In each iteration of training, we make sure to input the same number of units in the treatment and control groups to ensure that there exist the same number of factual outcomes as counterfactual outcomes in each head network to overcome the imbalanced classification. The inputs of discriminator are generated by the outcome generator $\Psi(R,T)$, and then the cross entropy loss of the counterfactual outcome discriminator is defined as:
\begin{equation}
\begin{split}
    \mathcal{L}_{\Phi,\Psi} = &-\frac{1}{2n}\sum_{t=0}^{1}\sum_{i=1}^{n}(p^{\text{truth}}_{ti}\log(p_{ti})+(1-p^{\text{truth}}_{ti})\log(1-p_{ti})),
    \label{Eqn: observed and potential outcome}
\end{split}
\end{equation}
where $p^{\text{truth}}_{ti}$ is the indicator that this input outcome for unit $i$ under treatment option $t$ is the observed factual outcome or inferred outcome from generator module, i.e., $p^{\text{truth}}_{ti}$ equals 1 or 0, separately. $P_{ti}$ is the probability judged by discriminator that how likely this input outcome for unit $i$ under treatment option $t$ is a factual outcome. 

Thus far, we have introduced the outcome generator to estimate potential outcomes for each unit across treatment and control groups, and the discriminator to determine if the potential outcome is factual, given a unit's confounder representation under treatment or control group. In the initial iterations of the model training, the outcome generator may generate potential outcomes that are very different from factual outcomes as determined by the discriminator. As the model is trained further, the discriminator may no longer be able to distinguish the generated counterfactual outcome and the factual outcome. At this point, we have attained all potential outcomes for each unit under treatment and control groups. For the training procedure of optimizing the outcome generator and discriminator, the minimax game is adopted. Putting all of the above together, the objective function of our Graph Infomax Adversarial Learning (GIAL) method is:

\begin{equation}
\begin{split}
    \text{min}_{\Psi} \text{max}_{\Phi,m} \ ( \mathcal{L}_\Psi + \alpha \mathcal{L}_m -\beta \mathcal{L}_{\Phi,\Psi}),
    \label{Eqn: minimax}
\end{split}
\end{equation}
where $\alpha$ and $\beta$ are the hyper-parameters controlling the trade-off among the outcome generator, mutual information, and discriminator.

\subsection{Overview of GIAL}

The proposed Graph Infomax Adversarial Learning method (GIAL) can estimate the treatment effects from networked observational data, which utilizes the graph neural network (GCN or GAT) and structure mutual information to learn the representations of hidden confounders and observed confounders, by mapping the feature covariates and network structure simultaneously into a representation space. Adversarial learning is also employed to mitigate the representation imbalance between treatment and control groups and to predict the counterfactual outcomes. After obtaining the counterfactual outcomes, GIAL can estimate the treatment effects. 

We summarize the procedures of GIAL as follows:
\begin{enumerate}

\item  Create the negative samples $(\Tilde{X}, A)$ by the row-wise shuffling of $X$ and keeping the original adjacency matrix $A$.

\item  Learn the representation space $R$ for the positive samples $(X,A)$ by function $g : X \times A \rightarrow R$ by a graph neural network.

\item  Learn the representation space $\Tilde{R}$ for the negative samples $(\Tilde{X},A)$ by function $g : \Tilde{X} \times A \rightarrow \Tilde{R}$ by the same graph neural network as Step 2.

\item  Utilize a structure summary function $f : R^{n\times d} \rightarrow S$ to summarize the learned representation into a graph-level structure representation, i.e., $s=f(g(X,A))$.

\item  Employ a discriminator $d: R \times S \rightarrow P$ to obtain $d(r_i,s)$ and $d(\Tilde{r}_i,s)$, which are the probabilities that the representations of $i$-th positive and negative samples are contained within the original graph structure summary $s$.

\item  Utilize functions $g$, $f$ and $d$ to maximize mutual information between $R$ and $S$.

\item  Use potential outcome generator $\Psi: R \times T \rightarrow Y$ to estimate the potential outcomes.

\item  Apply counterfactual discriminator $\Phi: R \times T \times (Y^f \text{or} \ \hat{Y}^{cf}) \rightarrow P$ to remove imbalance of confounder representations between treatment and control group.

\item Here, Steps 6, 7, and 8 in the procedure are jointly trained together by optimizing minimax rule Eq.~(\ref{Eqn: minimax}) about $\mathcal{L}_m$, $\mathcal{L}_\Psi$, and $\mathcal{L}_{\Phi,\Psi}$ to update parameters in $g$, $f$, $d$, $\Phi$, and $\Psi$.

\end{enumerate}

\section{Experiments}
In this section, we conduct experiments on two semi-synthetic networked datasets, including the BlogCatalog and Flickr, to evaluate the following aspects: (1) Our proposed method can improve treatment effect estimation with respect to average treatment effect and individualized treatment effect compared to the state-of-the-art methods. (2) The structure mutual information can help representations capture more hidden confounder information, and thus increase the predictive accuracy for counterfactual outcomes. (3) The proposed method is robust to the hyperparameters.

\subsection{Dataset}
\textbf{BlogCatalog.} BlogCatalog is a social blog directory that manages the bloggers and their blogs. In this dataset, each unit is a blogger and each edge represents the social relationship between two bloggers. The features are bag-of-words representations of keywords in bloggers' descriptions. We follow the assumptions and procedures of synthesizing the outcomes and treatment assignments in ~\cite{guo2019learning}. In this semi-synthetic networked dataset, the outcomes are the opinions of readers on each blogger and the treatment options are mobile devices or desktops on which blogs are read more. If the blogger's blogs are read more on mobile devices, the blogger is in the treatment group; if they are read more on desktops, the blogger is in the control group. We also assume that the topics of bloggers with the social relationship can causally affect their treatment assignment and readers' opinions on them. To model readers' preference on reading some topics from mobile devices and others from desktops, one LDA topic model ~\cite{guo2019learning} is trained.  Three settings of datasets are created with $k=0.5, 1,\text{and}\, 2$ that represent the magnitude of the confounding bias in the dataset. $k=0$ means the treatment assignment is random and there is no selection bias, and greater $k$ means larger selection bias. 

\textbf{Flickr.} Flickr is a popular photo-sharing and hosting service, and it supports an active community where people can share each other's photos. In the Flickr dataset, each unit is a user and each edge represents the social relationship between two users. The features of each user represent a list of tags of interest. The same settings and simulation procedures as BlogCatalog dataset are adopted here. Table~\ref{tab:freq} presents an overview of these two datasets.

\begin{table}
  \caption{Properties of BlogCatalog and Flickr datasets.}
  \label{tab:freq}
  \scalebox{0.9}{
  \begin{tabular}{lll}
    \toprule
    Datasets & \textbf{BlogCatalog}  & \textbf{Flickr}\\
    \midrule
    Nodes & 5,196  &  7,575\\
    Features & 8,189 & 12,047\\
    Edges & 171,743  & 239,738\\
    Treatments  &  2 & 2\\
  \bottomrule
\end{tabular}}
\end{table}

\subsection{Baseline Methods}

We compare the proposed GIAL with the following baseline methods.  \textbf{Network Deconfounder (ND)} ~\cite{guo2019learning} utilizes the GCN and integral probability metric to learn balanced representations to recognize patterns of hidden confounders from the network dataset. \textbf{Counterfactual Regression (CFRNET)} ~\cite{shalit2017estimating} maps the original features into a balanced representation space by minimizing integral probability metric between treatment and control representation spaces. \textbf{Treatment-agnostic Representation Networks (TARNet)} ~\cite{shalit2017estimating} is a variant of counterfactual regression without balance regularization. \textbf{Causal Effect Variational Autoencoder (CEVAE)} ~\cite{louizos2017causal} is based on Variational Autoencoder (VAE), which simultaneously estimates the unknown latent space summarizing the confounders and the causal effect. \textbf{Causal Forests (CF)} ~\cite{wager2018estimation} is a nonparametric forest-based method for estimating heterogeneous treatment effects by extending Breiman's random forest algorithm.  \textbf{Bayesian Additive Regression Trees (BART)}~\cite{chipman2010bart} is a nonparametric Bayesian regression model, which uses dimensionally adaptive random basis elements.

\subsection{Descriptive Data Analysis}
\begin{figure}[h!]
    \centering
    \includegraphics[width=0.2\textwidth]{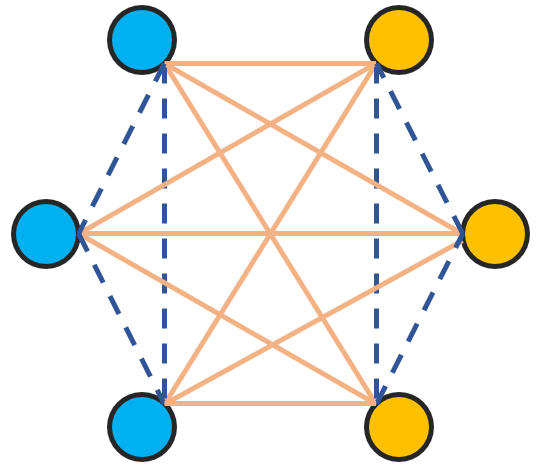}
    \vspace{-2mm}
    \caption{Example of complete graph. The solid line represents heterogeneous edge and the dashed line means homogeneous edge.}
    \vspace{-2mm}
    \label{complete}
\end{figure}

\begin{table}
  \caption{Summary of homogeneous edges and heterogeneous edges for the BlogCatalog datasets and  Flickr datasets.}
  \vspace{-2mm}
  \label{homogeneous}
  \centering
  \scalebox{0.9}{
  \begin{tabular}{llll}
    \toprule
    \multicolumn{1}{c}{Dataset} & \multicolumn{1}{c}{k} & \multicolumn{1}{c}{Homogeneous} & \multicolumn{1}{c}{Heterogeneous}                  \\
    \midrule
    & 0.5 & \textbf{94524.5} & 77218.5\\
   BlogCatalog & 1 & \textbf{101102.8} & 70640.2\\
    & 2 & \textbf{116031.8} & 55711.2\\
    \midrule
    & 0.5 & \textbf{124320.9} & 115417.1\\
   Flickr & 1 & \textbf{130978.5} & 108759.5\\
    & 2 & \textbf{141957.3} & 97780.7\\

    \bottomrule
  \end{tabular}}
  \vspace{-2mm}
\end{table}

Before estimating the treatment effects from these two networked datasets, we provide the descriptive data analysis to demonstrate the existence of network structural imbalance in the networked data for causal inference problems.

\begin{table*}[t]
  \caption{Performance comparison on BlogCatalog and Flickr datasets with different $k\in{0.5, 1, 2}$. We present the mean value of $\sqrt{\epsilon_\text{PEHE}}$ and $\epsilon_\text{ATE}$ on the test sets. Results of baseline methods on the same datasets are reported in~\cite{guo2019learning}.}
  \vspace{-2mm}
  \label{result}
  \centering
  \scalebox{0.9}{
  \begin{tabular}{lllllllllllll}
    \toprule
     & \multicolumn{6}{c}{BlogCatalog} & \multicolumn{6}{c}{Flickr} \\
    \cmidrule(lr){2-7} \cmidrule(lr){8-13} 
    
     & \multicolumn{2}{c}{k=0.5} & \multicolumn{2}{c}{k=1}  & \multicolumn{2}{c}{k=2}  & \multicolumn{2}{c}{k=0.5} & \multicolumn{2}{c}{k=1}  & \multicolumn{2}{c}{k=2}                 \\
    \cmidrule(lr){2-3} \cmidrule(lr){4-5} \cmidrule(lr){6-7} \cmidrule(lr){8-9} \cmidrule(lr){10-11} \cmidrule(lr){12-13}
    Method     & $\sqrt{\epsilon_\text{PEHE}}$   & $\epsilon_\text{ATE}$  & $\sqrt{\epsilon_\text{PEHE}}$     & $\epsilon_\text{ATE}$ & 
    $\sqrt{\epsilon_\text{PEHE}}$     & $\epsilon_\text{ATE}$ & $\sqrt{\epsilon_\text{PEHE}}$   & $\epsilon_\text{ATE}$  & $\sqrt{\epsilon_\text{PEHE}}$     & $\epsilon_\text{ATE}$ & 
    $\sqrt{\epsilon_\text{PEHE}}$     & $\epsilon_\text{ATE}$\\
    \midrule
     
    BART~\cite{chipman2010bart} & 4.808&2.680 &5.770 &2.278 &11.608 &6.418 & 4.907 & 2.323 & 9.517 & 6.548 & 13.155 & 9.643\\
    CF~\cite{wager2018estimation} &7.456 &1.261 &7.805 &1.763 &19.271 &4.050 & 8.104& 1.359  & 14.636& 3.545 & 26.702 & 4.324\\
    CEVAE~\cite{louizos2017causal} & 7.481& 1.279&10.387 &1.998 & 24.215&5.566 & 12.099  &1.732 & 22.496& 4.415&42.985 & 5.393\\
    TARNet~\cite{shalit2017estimating} & 11.570& 4.228& 13.561& 8.170& 34.420&13.122 & 14.329  &3.389 & 28.466& 5.978& 55.066& 13.105\\
    $\text{CFRNET}_\text{MMD}$~\cite{shalit2017estimating} &11.536 &4.127 & 12.332&5.345 & 34.654& 13.785 &13.539 &3.350 & 27.679& 5.416& 53.863& 12.115\\
    $\text{CFRNET}_\text{Wass}$~\cite{shalit2017estimating} & 10.904& 4.257& 11.644&5.107 & 34.848&13.053 & 13.846&3.507 &27.514 & 5.192& 53.454& 13.269\\
    ND~\cite{guo2019learning} & 4.532  &0.979 & 4.597&0.984 & 9.532&2.130 &4.286 & 0.805&5.789 & 1.359& 9.817& 2.700\\

    \midrule
    $\text{GIAL}_\text{GAT}$ (Ours) &4.215 &0.912 &4.258 &0.937 &9.119 &1.982 &4.015 &0.773 &5.432 & 1.2312&9.428&2.586\\
    $\text{GIAL}_\text{GCN}$ (Ours) & \textbf{4.023} &\textbf{0.841} &\textbf{4.091} &\textbf{0.883} &\textbf{8.927} &\textbf{1.780} &\textbf{3.938} &\textbf{0.682}& \textbf{5.317}&\textbf{1.194} &\textbf{9.275} &\textbf{2.245}\\

    \bottomrule
  \end{tabular}}
  \vspace{-2mm}
\end{table*}

\begin{table}[t]
   \vspace{-2mm}
  \caption{Summary of results in ablation studies.}
  \vspace{-2mm}
  \label{ablation}
  \centering
  \scalebox{0.9}{
  \begin{tabular}{lllllll}
    \toprule
     & \multicolumn{2}{c}{k=0.5} & \multicolumn{2}{c}{k=1}  & \multicolumn{2}{c}{k=2}                 \\
    \cmidrule(lr){2-3} \cmidrule(lr){4-5} \cmidrule(lr){6-7} 
         & $\sqrt{\epsilon_\text{PEHE}}$   & $\epsilon_\text{ATE}$  & $\sqrt{\epsilon_\text{PEHE}}$     & $\epsilon_\text{ATE}$ & 
    $\sqrt{\epsilon_\text{PEHE}}$     & $\epsilon_\text{ATE}$ \\
      \midrule
    \textbf{BlogCatalog}  & \multicolumn{6}{c}{}  \\

    GIAL &4.023 &0.841 &4.091 &0.883 &8.927 &1.780 \\
    GIAL (w/o SMI) &4.422&0.982 &4.481 &0.981 &9.315 &2.142\\
    GIAL (w/o CD) &4.482 & 0.987&4.951 &1.023 &13.598 &3.215\\
    
    \midrule
    \textbf{Flickr} & \multicolumn{6}{c}{}  \\
    
    GIAL &3.938 &0.682& 5.317&1.194 &9.275 &2.245\\
    GIAL (w/o SMI) &4.158 &0.792 &5.694 &1.375 & 9.673&2.661 \\
    GIAL (w/o CD) &4.284 &0.812 &6.127 &1.435 &11.524 &3.564 \\
    \bottomrule
  \end{tabular}
  }
  \vspace{-2mm}
\end{table}

According to graph theory, in the complete graph which is a simple undirected graph where every pair of distinct nodes is connected by a unique edge, there are $\frac{n\times(n-1)}{2}$ edges for $n$ nodes. We assume that the $n$ nodes are evenly divided into treatment group and control group with the same $\frac{n}{2}$ nodes in each group, and also each node has the same possibility to have an edge (relationship) with another node regardless of the node's treatment assignment. Then, this graph is still a complete graph with $\frac{n\times(n-1)}{2}$ edges. Now the edges in this graph are put into two categories: (a) the homogeneous group including the edges that link the nodes in the same group (treatment-treatment or control-control); (b) the heterogeneous group including the edges that link the nodes in different groups (treatment-control). Under the assumption that each node has the same possibility to be connected with another node regardless of the node's treatment assignment, we can find that in the homogeneous group, there are $\frac{n^2}{4}-\frac{n}{2}$ edges and in the heterogeneous group, there are $\frac{n^2}{4}$ edges. The number of edges in the heterogeneous group should be greater than that in the homogeneous group. For example, as shown in Fig.~\ref{complete}, there is one complete graph with 6 nodes including 3 treatment nodes and 3 control nodes. The heterogeneous group has 9 edges, while the homogeneous group has 6 edges.

We separately calculate the average numbers of homogeneous edges and heterogeneous edges for the BlogCatalog datasets and Flickr datasets, then report them in Table~\ref{homogeneous}. We can observe that the homogeneous edges are consistently greater than the heterogeneous edges for both datasets with different $k$. This result totally agrees with our expectation that, in the causal inference problem, the network structure is imbalanced. Therefore, the relationship is more likely to appear among people who are in the same group. This is the major difference between traditional graph learning tasks and the causal inference task on networked data, which is also the motivation of our proposed model.

\subsection{Experimental Settings}
In the following experiments, we randomly sample $60\%$ and $20\%$ of the units as the training set and validation set, and use the remaining $20\%$ units to form the test set. For each dataset with a different imbalance $k$, the simulation procedures are repeated 10 times and we report the average mean.

\textbf{GIAL.} By using different graph neural networks to learn the representation space from the networked dataset, the proposed GIAL method has two variants denoted as GIAL$_\text{GCN}$ and GIAL$_\text{GAT}$, which adopt the original implementation of graph convolutional network ~\cite{kipf2016semi} and graph attention network (GAT)~\cite{velickovic2019deep}, respectively. Besides, a squared $l_2$ norm regularization with hyperparameter $10^{-4}$ is added into our model to mitigate the overfitting issue. The hyperparameters of our method are chosen based on performance on the validation dataset, and the searching range is shown in Table~\ref{hyperparameter}. The Adam SGD optimizer ~\cite{kingma2014adam} is used to train the final objective function Eq.~(\ref{Eqn: minimax}) with an initial learning rate of 0.001 and an early stopping strategy with patience of 100 epochs.

\begin{table}[th!]
  \vspace{-2mm}
  \caption{Hyperparameters and ranges.}
  \vspace{-2mm}
  \label{hyperparameter}
  \centering
  \scalebox{0.9}{
  \begin{tabular}{ll}
    \toprule
    \multicolumn{1}{c}{Hyperparameter} & \multicolumn{1}{c}{Range}  \\
     \midrule
     $\alpha$,  $\beta$ & 0, $10^{-4}$,$10^{-3}$ ,$10^{-2}$ ,$10^{-1}$ \\
     \midrule
    Dim. of confounder representation & 50, 100, 150, 200\\
    \midrule
    No. of GCN and GAT layers & 1, 2, 3\\
    \midrule
    No. of attention heads in GAT & 1, 2, 3, 4\\
    \midrule
    No. of outcome generator layer & 1, 2, 3, 4\\
    \bottomrule
  \end{tabular}
  }
\end{table}

\textbf{Baseline Methods.} BART, CF, CEVAE, TARNet, and CFRNET are not originally designed for the networked observational data, so they cannot directly utilize the network information. To be fair, we concatenate the corresponding row of adjacency matrix to the original features, but this strategy cannot effectively improve the performance of baselines due to the curse of dimensionality. Besides, we adopt their default hyperparameter settings~\cite{guo2019learning}. 

\subsection{Results}
For the BlogCatalog and Flickr datasets, we adopt two commonly used evaluation metrics to evaluate the performance of our method and baselines. The first one is the error of ATE estimation, which is defined as $\epsilon_\text{ATE}  = |\text{ATE} - \widehat{\text{ATE}}|$, where \text{ATE} is the true value and $\widehat{\text{ATE}}$ is an estimated \text{ATE}. The second one is the error of expected precision in estimation of heterogeneous effect (PEHE)~\cite{hill2011bayesian}, which is defined as $\epsilon_\text{PEHE}  = \frac{1}{n}\sum_{i=1}^{n}(\text{ITE}_i-\widehat{\text{ITE}}_i)^2$, where $\text{ITE}_i$ is the true \text{ITE} for unit $i$ and $\widehat{\text{ITE}}_i$ is an estimated \text{ITE} for unit $i$.

Table~\ref{result} shows the performance of our method and baseline methods on the BlogCatalog and Flickr datasets over 10 realizations. We report the average results of $\sqrt{\epsilon_\text{PEHE}}$ and $\epsilon_\text{ATE}$ on the test sets. $\text{GIAL}_\text{GCN}$ achieves the best performance with respect to $\sqrt{\epsilon_\text{PEHE}}$ and $\epsilon_\text{ATE}$ in all cases of both datasets. Although the $\text{GIAL}_\text{GAT}$ also has obvious improvements compared to baseline methods, it is outperformed by $\text{GIAL}_\text{GCN}$. GCN demonstrates clear superiority over GAT when recognizing patterns of hidden confounders from imbalanced network structure. Because $k=0.5, 1,\text{and}\, 2$ is used to represent the magnitude of the confounding bias in both datasets, results show that GIAL consistently outperforms the baseline methods under different levels of divergence, and our method is robust to a high level of confounding bias. Compared to baseline methods (e.g., CFRNET) only relying on observed confounders but without utilizing the network information, our model is capable of recognizing the patterns of hidden confounders from the network structure. Compared to baseline methods with learning network information (e.g., ND), our model has significant performance advantages, which demonstrates our model can capture more information from an imbalanced network structure. The reason is that our method maximizes the structure mutual information, instead of directly adopting the graph learning method without considering the specificity of networked data in the causal inference problem.

\subsection{Model Evaluation}
Experimental results on both datasets show that GIAL obtains a more accurate estimation of the ATE and ITE than the state-of-the-art methods. We further evaluate the performance of GIAL from two perspectives, including the effectiveness of each component, and its robustness to hyper-parameters.

We perform two ablation studies of $\text{GIAL}_\text{GCN}$ on both datasets. The first one is GIAL (w/o SMI) where the structure mutual information maximizing module is removed. We directly adopt graph neural networks to learn the representation space without considering the structural imbalance of networked data. The second ablation study is GIAL (w/o CD) where the counterfactual outcome discriminator is removed and there is not any restriction on the divergence between the representation distributions of treatment and control groups. 

As shown in Table~\ref{ablation}, the performance becomes poor after removing either the structure mutual information or counterfactual outcome discriminator, compared to the original GIAL. More specifically, after removing the structure mutual information, $\sqrt{\epsilon_\text{PEHE}}$ and $\epsilon_\text{ATE}$ increase dramatically and have similar performance to other baseline methods. Besides, as the bias ($k$) increases, the difference between the performance of GIAL (w/o CD) and the original GIAL increases further. Therefore, the structure mutual information and counterfactual outcome discriminator are essential components of our model.

\begin{figure*}[t]
    \centering
    \includegraphics[width=1\linewidth]{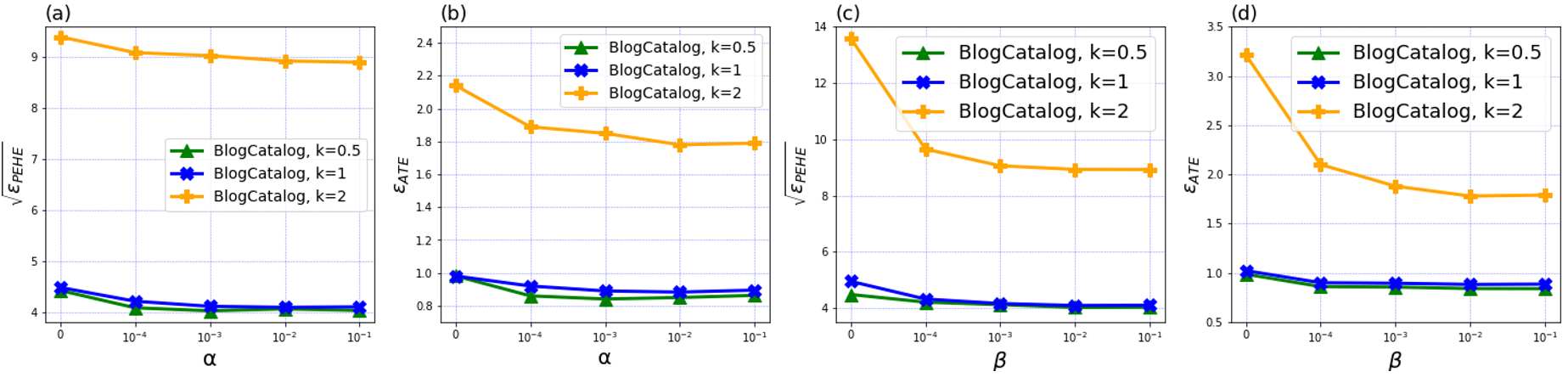}
    \caption{Sensitivity analysis for $\alpha$ and $\beta$ of structure mutual information and counterfactual outcome discriminator.}
    \label{sensitivity}
\end{figure*}

Next, we explore the model's sensitivity to the most important parameters $\alpha$ and $\beta$, which control the ability to capture the graph structure and handle the confounding bias when estimating the potential outcomes. We show the results of $\sqrt{\epsilon_\text{PEHE}}$ and $\epsilon_\text{ATE}$ on BlogCatalog dataset with different $k$ in Fig.~\ref{sensitivity}. We observe that the performance is stable over a large parameter range. It confirms the effectiveness and robustness of structure mutual information and counterfactual outcome discriminator in GIAL, which is consistent with our ablation studies, i.e., GIAL (w/o SMI) and GIAL (w/o CD).

\section{Related Work}
The related work is presented along with two directions: learning causal effects from observational data and graph neural networks.

Various causal effect estimation methods for observational data have sprung up. For most existing methods,  the strong ignorability assumption is the most important prerequisite. However, this assumption might be untenable in practice. A series of methods have been proposed to relax the strong ignorability assumption. A latent variable is inferred as a substitute for unobserved confounders~\cite{wang2019blessings}. Variational Autoencoder has been used to infer the relationships between the observed confounders based on the assumption joint distribution of the latent confounders and the observed confounders can be approximately recovered solely from the observations ~\cite{louizos2017causal}. Recently, some work aims to relax the strong ignorability assumption via network knowledge, where the network connecting the units is a proxy is for the unobserved confounding. The network deconfounder~\cite{guo2019learning} learns representations of confounders from network data by adopting the graph convolutional networks. Another work utilizes graph attention networks to learn representations and mitigates confounding bias by representation balancing and treatment prediction, simultaneously ~\cite{guo2020ignite}. Causal network embedding (CNE)~\cite{veitch2019using} is proposed to learn node embeddings from network data to represent confounders by reducing the causal estimation problem to a semi-supervised prediction of both the treatments and outcomes. For the existing methods about networked data, they do not dig deeply on what is the essential difference between the networked data under the causal inference problem and the networked data for traditional graph learning tasks such as node classification, link detection, etc. This is the reason why we propose this GIAL model, instead of directly adopting the GCN or GAT to learn the representation from the networked data.

Graph learning is increasingly becoming fascinating as more and more real-world data can be modeled as networked data. Graph convolutional network~\cite{kipf2016semi} is an effective approach for semi-supervised learning on networked data, via a localized first-order approximation of spectral graph convolutions. Graph attention network (GAT) ~\cite{velivckovic2017graph} is an attention-based architecture leveraging masked self-attentional layers where nodes are able to attend over their neighborhoods' features. Deep graph infomax (DGI) ~\cite{velickovic2019deep} is one approach for learning node representations within networked data in an unsupervised manner, which relies on maximizing mutual information between patch representations and high-level summaries of graphs. In our model, we extend the idea in DGI originally aimed for unsupervised learning to representation learning under the causal inference setting. Utilizing the structure mutual information can help representations capture the imbalanced structure that is specific to the causal inference problem.

\section{Conclusion}

In this paper, we propose the Graph Infomax Adversarial Learning method (GIAL) to capture the hidden confounders and estimate the treatment effects from networked observational data. GIAL makes full use of the network structure to capture more information by recognizing the imbalance in the network structure. Our work clarifies the greatest particularity of networked data under the causal inference problem compared with traditional graph learning tasks, that is, the structural imbalance due to confounding bias between treatment and control groups. Extensive experiments show the effectiveness and advantages of the proposed GIAL method.
 
\section*{ACKNOWLEDGMENTS}
We would like to thank the anonymous reviewers for their insightful comments. This research is supported in part by the U.S. Army Research Office Award under Grant Number W911NF-21-1-0109. 

\bibliographystyle{ACM-Reference-Format}

\end{document}